\documentclass[letterpaper, 10 pt, conference]{ieeeconf}
\IEEEoverridecommandlockouts
\usepackage[utf8]{inputenc}
\usepackage{amsmath,amssymb,amsfonts}
\usepackage{blindtext, graphicx}
\PassOptionsToPackage{greek,english}{babel}
\usepackage{textcomp}
\usepackage{xcolor}
\usepackage{booktabs} % For formal tables
\usepackage{bbding}
\usepackage{soul,color}
\usepackage{comment} 
\usepackage{kantlipsum}
\usepackage{multicol}
\usepackage{microtype}
\usepackage{array}
\usepackage{tabulary}
\usepackage{chngcntr}
\usepackage{multirow}
\usepackage[T1]{fontenc}
\usepackage{amssymb}

\usepackage{tablefootnote}
\usepackage{threeparttable}

\usepackage{cleveref}

\let\Setlength\setlength % <--- Important: place this line before loading calc
\usepackage{calc}
\newlength{\arrayrulewidthOriginal}

\DeclareMathAlphabet\mathzapf{T1}{pzc}{mb}{it}
\usepackage[style=ieee,minbibnames=1,maxbibnames=2,doi=false,isbn=false,url=false,eprint=false,date=year,bibencoding=utf8]{biblatex}
\AtEveryBibitem{\clearfield{pages}}
\AtEveryBibitem{\clearfield{volume}}
\AtEveryBibitem{\clearfield{number}}
\AtEveryBibitem{\clearfield{note}}
\AtEveryBibitem{\clearlist{language}}

\newcolumntype{K}[1]{>{\centering\arraybackslash}p{#1}}
\usepackage{float}
\usepackage[linesnumbered,ruled,vlined]{algorithm2e}
\usepackage{bm}

\newcolumntype{M}[1]{>{\centering\arraybackslash}m{#1}}
\Setlength{\intextsep}{5pt}
\renewcommand{\baselinestretch}{1} 

\IEEEoverridecommandlockouts                              % This command is only needed if 
\begin{document}
\title{\textbf{Exploration of Hyperdimensional Computing Strategies \\ for Enhanced Learning on Epileptic Seizure Detection}
\thanks{This work has been partially supported by the ML-Edge Swiss National Science Foundation (NSF) Research project (GA No. 200020182009/1), and the PEDESITE Swiss NSF Sinergia project (GA No. SCRSII5 193813/1).}
\thanks{$^{1}$ U. Pale, T. Teijeiro, and D. Atienza are with the Embedded Systems Laboratory (ESL) of Swiss Federal Institute of Technology Lausanne (EPFL), Switzerland.\newline
{\tt\footnotesize \{una.pale, tomas.teijeiro, david.atienza\}@epfl.ch}}
}

%\author{ \parbox{3 in}{\centering Huibert Kwakernaak*
%         \thanks{*Use the $\backslash$thanks command to put information here}\\
%         Faculty of Electrical Engineering, Mathematics and Computer Science\\
%         University of Twente\\
%         7500 AE Enschede, The Netherlands\\
%         {\tt\small h.kwakernaak@autsubmit.com}}
%         \hspace*{ 0.5 in}
%         \parbox{3 in}{ \centering Pradeep Misra**
%         \thanks{**The footnote marks may be inserted manually}\\
%        Department of Electrical Engineering \\
%         Wright State University\\
%         Dayton, OH 45435, USA\\
%         {\tt\small pmisra@cs.wright.edu}}
%}

\author{ Una Pale, Tomas Teijeiro, and David Atienza$^{1}$ }

\maketitle

\begin{abstract}
Wearable and unobtrusive monitoring and prediction of epileptic seizures has the potential to significantly increase the life quality of patients, but is still an unreached goal due to challenges of real-time detection and wearable devices design. 
Hyperdimensional (HD) computing has evolved in recent years as a new promising machine learning approach, especially when talking about wearable applications. 
But in the case of epilepsy detection, standard HD computing is not performing at the level of other state-of-the-art algorithms. This could be due to the inherent complexity of the seizures and their signatures in different biosignals, such as the electroencephalogram (EEG), the highly personalized nature, and the disbalance of seizure and non-seizure instances. %, and scarcity of labeled data.
In the literature, different strategies for improved learning of HD computing have been proposed, such as iterative (multi-pass) learning, multi-centroid learning and learning with sample weight ("OnlineHD"). Yet, most of them have not been tested on the challenging task of epileptic seizure detection, and it stays unclear whether they can increase the HD computing performance to the level of the current state-of-the-art algorithms, such as random forests. 
Thus, in this paper, we implement different learning strategies and assess their performance on an individual basis, or in combination, regarding detection performance and memory and computational requirements. Results show that the best-performing algorithm, which is a combination of multi-centroid and multi-pass, can indeed reach the performance of the random forest model on a highly unbalanced dataset imitating a real-life epileptic seizure detection application. %These results show possible directions for further improvements of HD computing on complex classification problems. 

\end{abstract}

% \begin{IEEEkeywords}
% Hyperdimensional computing, Epilepsy, Seizure detection, EEG, IEEG, Wearable devices, Low-power signal processing.
% \end{IEEEkeywords}

\section{Introduction}
\bstctlcite{IEEEexample:BSTcontrol}

Hyperdimensional (HD) computing is a novel machine learning paradigm inspired by neuroscience that has attracted lots of interest in the last years in many different domains. For example, in the biomedical domain it has been used for emotion recognition from GSR (galvanic-skin response), electrocardiogram (ECG) and electroencephalogram (EEG)~\cite{chang_hyperdimensional_2019}, EEG error-related potentials detection~\cite{rahimi_hyperdimensional_2020}, electromyogram (EMG), gesture recognition~\cite{rahimi_hyperdimensional_2016}, epileptic seizure detection from EEG~\cite{asgarinejad_detection_2020}, etc. 
HD computing is an interesting alternative to standard machine learning approaches and it offers opportunities for continuous online learning~\cite{moin_wearable_2021,benatti_online_2019}, semi-supervised~\cite{imani_semihd_2019}, distributed~\cite{imani_framework_2019} or multi-centroid learning~\cite{pale_multi-centroid_2021}.
Furthermore, it is highly efficient due to its lower energy and memory requirements~\cite{burrello_ensemble_2021,asgarinejad_detection_2020, imani_sparsehd_2019}, and offers opportunities for design novel algorithms~\cite{imani_semihd_2019,hernandez-cano_real-time_2021, pale_multi-centroid_2021}, which makes it interesting for wearable devices and applications. Thus, hardware implementations and optimizations for HD computing are a common topic of recent works, and show promising results~\cite{salamat_f5-hd_2019, gupta_felix_2018}. 

In recent years, a lot of effort has been put into designing wearable devices for patient monitoring, with detection and prediction capabilities. Epilepsy monitoring and real-time seizures detection are one of such applications.
Epilepsy is a chronic neurological disorder characterized by the unpredictable occurrence of seizures, affecting a significant portion of the world population (0.6 to 0.8\%)~\cite{mormann_seizure_2007}. It attracts a lot of medical attention as, despite pharmacological treatments, one-third of patients still suffer from seizures~\cite{schmidt_evidence-based_2012}. The unexpected occurrence of seizures imposes serious health risks and many restrictions on daily life. As such, solutions that would allow continuous unobtrusive monitoring and a reliable detection (and ideally prediction) of seizures will be of high importance. Such technology will also be instrumental in deepening knowledge and designing novel treatments.

\begin{figure*}[]
    \centering
    %\vspace{2mm}
    \includegraphics[width=0.95\linewidth]{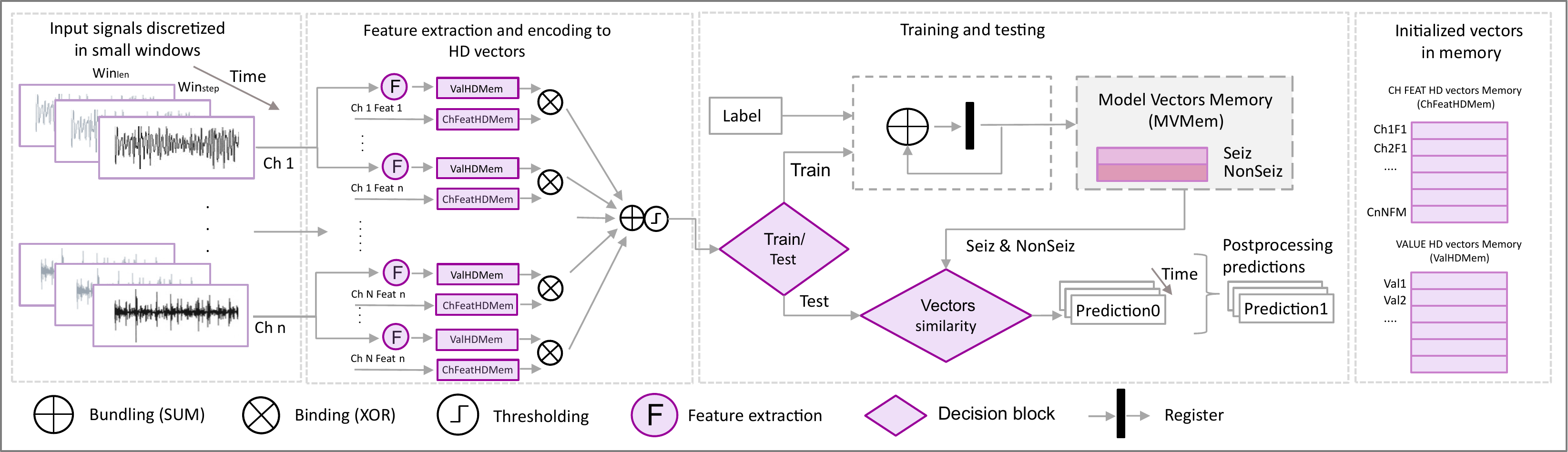}
    \vspace*{-2mm}
    \caption{\small{Schematic of the HD computation workflow. Adapted from~\cite{pale_multi-centroid_2021}. }} 
    \label{fig:workflowSchematic}
    \vspace{-4mm}
\end{figure*}

Epilepsy is not only challenging from a medical perspective related to its origins, treatment, and prevention but also from the engineering perspective of detection from the recorded physiological signals. Even from multi-channel EEG recordings, it is a big challenge to design a lightweight, real-time detection system with (almost) no false positive alarms. %, which is important for the user-acceptance of such systems. 
Reasons are multi-fold: huge disbalance in data distribution (i.e., the amount of seizure vs. non-seizure data), scarcity of training data available, highly variable and personalized morphologies of signals, as well as computational and memory requirements of algorithms used. 

HD computing comes as an interesting option due to the aforementioned positive aspects. Despite initial optimistic performance reported in the literature~\cite{burrello_ensemble_2021,asgarinejad_detection_2020}, HD computing is not performing so well when taking into account more realistic data distributions~\cite{pale_multi-centroid_2021}. As will be shown in this paper, when using ten times more non-seizure data, performance is significantly reduced with the HD computing approach when compared to the Random Forest model.

At the same time, various strategies have been proposed in the literature to improve learning and detection with HD computing. For example, the iterative learning approach~\cite{imani_adapthd_2019} has been proposed instead of single-pass learning, but it has not been fully explored yet for epilepsy. Further, the multi-centroid approach~\cite{pale_multi-centroid_2021}, which is appropriate for the high variability of seizure and non-seizure signals, also showed significant improvements when compared to the standard single-centroid approach. In the end, the "OnlineHD" approach~\cite{hernandez-cano_real-time_2021} also works in a single-pass manner but multiplies vectors with a weight factor depending on the novelty each data window brings. It aims, as well as both iterative and multi-centroid approaches, to overcome the dominance problem of more common patterns in the final prototype vectors. This approach also has not yet been tested for the challenging task of epileptic seizure detection. 

Thus, following the challenges epileptic seizure detection poses, we aim to test and compare various algorithms for HD computing improvement on this problem, hoping to achieve a performance that would be acceptable by users. In this work, we contribute to the state of the art in the following manner:
\begin{itemize}
    \item We systematically compare the performance of the standard single-pass one-centroid HD computing approach with the classical random forest learning model on epileptic seizure detection with more realistic data distributions and show that HD computing still has a performance gap to cross. 
    \item We then implement several existing proposals for improved HD computing learning: iterative (multi-pass) one-centroid learning, multi-centroid single-pass as well as "OnlineHD". Not all of them have been yet tested for epileptic seizure detection. 
    \item We combine two strategies to test whether multi-centroid with multi-pass approach can bring additional improvements to the performance. 
    \item Finally, we compare the mentioned strategies in terms of their performance and their memory and computational requirements with a wearable implementation in mind. 
\end{itemize}

\section{HD computing learning strategies}

\subsection{Traditional HD computing workflow}
HD computing is based on computations with very long and redundant (mostly binary) vectors (usually >10000 dimensions), which represent information in a condensed and distributed way. %It as inspired by the neuroscience research hypothesis that the brain's computation is based on the high-dimensional randomized representation of data rather than scalar numerical values~\cite{kanerva_hyperdimensional_2009}.
Calculating and learning with vectors is based on a few specific algebraic properties. The two most important ones are: 1) any randomly chosen pair of vectors are nearly orthogonal and 2) when summing two or more vectors, the result will be with high probability more similar to the added vectors than to any other randomly chosen vector. These properties are crucial for learning and inference. 

\begin{figure*}[]
    \centering
    %\vspace{2mm}
    \includegraphics[width=0.98\linewidth]{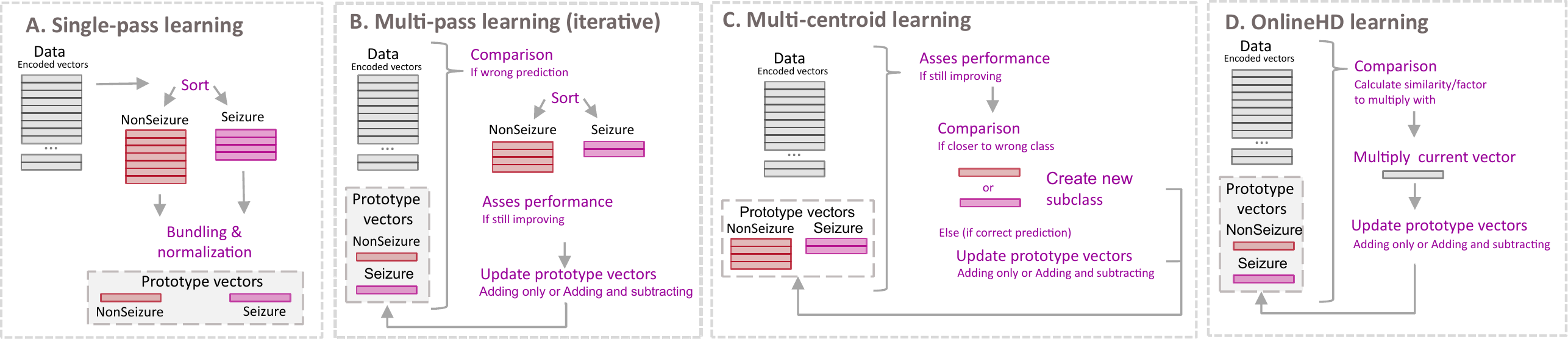}
    \vspace*{-2mm}
    \caption{\small{Schematic of different improvements on HD computing learning: A. Standard single-pass single-centroid HD learning, B. Multi-pass (iterative) single-centroid learning, C. Multi-centroid single-pass learning and D. OnlineHD single-pass learning. }} 
    \label{fig:diffApproachesSchematic}
    \vspace{-4mm}
\end{figure*}

More specifically, HD learning starts by encoding data and its relations to HD vectors. As illustrated in Fig.~\ref{fig:workflowSchematic}, baseline vectors representing different scalar values, features and channels are combined during the encoding stage to get one vector. This vector represents that specific data sample (window), instead of a feature set, as in other ML approaches. The training phase is relatively simple and consists of summing (bundling) all vectors from the same class to one prototype vector representing each class. Summation of the vectors is usually done by bit-wise summation, followed by majority voting normalization. 
In the end, for inference, a vector representing the current data sample is compared with prototype vectors of all classes, and the label of the most similar one is given as output. The most common measure of similarity is the Hamming distance in the case of binary HD vectors (elements are 0 or 1), but cosine or dot products can also be used in the case of integer or float HD vectors. 

Traditionally, HD computing classifiers have been based on a single-pass learning and a single-centroid model vector per class as an outcome of learning, as illustrated in a simplified way in Fig.~\ref{fig:diffApproachesSchematic}A. However, many current machine learning problems are challenging due to the immense complexity and variability of patterns in data, especially when compared to the amount of training data available. 
Epilepsy is one such case, where electroencephalogram (EEG) signatures of epileptic seizures are highly unique and variable among people, brain states, and time instances, especially if they are grouped under only two given labels (seizure and non-seizure). Non-seizure data represents many different brain states, such as awake, sleeping, physical or mental effort conditions, which can significantly increase data variability grouped under one class. But, all of these states have their own brain signatures that need to be learned. 
In the standard single-pass one-centroid HD approach, all data samples are equally important during learning, leading to more common patterns dominating the prototype vectors. This means that less common patterns could be potentially under-represented and wrongly predicted even on the same training data. 

Thus, in this work, we explore and compare different proposed strategies that can tackle this problem: iterative (multi-pass) learning, multi-centroid, multi-centroid with iterative learning, and "OnlineHD". We find epileptic seizure detection a perfect test case for this due to the variability of patterns in both seizure and non-seizure classes, the inherent disbalance in the amount of seizure/non-seizure recordings, and a generally relatively small size of training data.

\subsection{Multi-pass learning approach}
%\vspace{-2mm}
The first approach aimed at overcoming the problem of dominance of common patterns is iterative learning~\cite{imani_adapthd_2019}. We call it also multi-pass learning to clarify its relation to other approaches. In the first pass of learning, all samples are added to the corresponding data class, the same as in traditional single-pass learning. Next, an iterative process of multiple passes starts wherein each pass for each sample prediction is given based on the learned prototype vectors from the previous pass. In case of a wrong prediction, this sample is added again to the correct class. In this way, less common patterns that got under-represented are strengthened by adding them multiple times. After each pass, performance is evaluated on the same training set. The multi-pass process stops once there is no more significant improvement in performance. The workflow of iterative learning is shown in Fig.~\ref{fig:diffApproachesSchematic}B. 

This approach can be improved if the vector of a wrongly classified sample is added to the correct class prototype vector and also subtracted from the other ones. Further, before adding/subtracting vectors to prototype vectors, it can be multiplied with a factor (learning rate) to increase its weight. In~\cite{imani_adapthd_2019}, authors test different values of multiplication factors (learning rate) and their influence on the number of iterations needed to achieve a stable performance and the final performance itself. They show that small factors lead to clear performance increase but require many iterations. On the other side, too big multiplication factors lead to fluctuations in performance, thus potentially never converging and never finishing the training procedure. Hence, they also tested an adaptable threshold strategy to exploit the advantages of a large learning rate at the beginning and a small learning rate in the end for fine-tuning the performance. %The authors proposed and tested two methods for adaptive retraining: iteration-dependent and data-dependent learning rate and their combination. The iteration-dependant approach measures the classification accuracy during retraining and changes the learning rate based on that, whereas the data-dependent approach changes the learning rate for each data point based on how far off the data was misclassified.
They tested these approaches on four different classification applications and showed performance improvement as well as improved energy and computational complexity. 

Unfortunately, none of the applications used for testing was epileptic seizure detection, and one could argue that the chosen problems were less complex and challenging than the epileptic seizure one. Further, in order to be able to compare properly with other approaches (explained below), we implemented this approach as well.  

% \begin{figure*}[]
%     \centering
%     \vspace{2mm}
% \includegraphics[width=\linewidth]{Images/PerformanceMetricsIllustration.pdf}
%     \caption{\small{Illustration of performance metrics on the level of episodes and duration of seizure. Adapted from~\cite{pale_multi-centroid_2021}.}} 
%     \label{fig:perfMetricsIllustration}
%     \vspace{-4mm}
% \end{figure*} 

\subsection{Multi-centroid learning approach}
%MULTI CENTROID 
Another way to not let the frequency of certain patterns make them dominate the model, leading to under-representation and worse classification of less common patterns, is to treat them as separate sub-classes (centroids) of the same class. In this way, they are not all accumulated to the same prototype vectors, but it is possible to have multiple vectors representing the same class. This approach was explored in~\cite{pale_multi-centroid_2021} and has shown to significantly improve epileptic seizure detection. Balanced datasets, with the same amount of seizure and non-seizure data, are commonly used, but are not realistic in real-life applications. Thus, authors compared performance also on unbalanced datasets with more non-seizure than seizure data. The more unbalanced the dataset was, the bigger the performance improvement when compared to the single-centroid approach. 

The workflow of the approach is illustrated in Fig.~\ref{fig:diffApproachesSchematic}C. This approach is interesting as it allows the semi-supervised creation of an unlimited number of centroids (sub-classes) of each class. Namely, if the current sample vector is more similar to the wrong class than any correct sub-class (meaning that it would be wrongly classified), a new subclass of the correct class is created. This process can sometimes lead to many sub-classes where some have only a few samples that contributed (were added) to it. Hence, in the last step of the workflow, the number of sub-classes is reduced by either removing the less common sub-classes or clustering them with the most similar sub-classes sharing the same label. This step significantly reduces the number of sub-classes and thus the memory required to store them while still achieving higher performance by allowing several centroids. 
When compared to the iterative multi-pass approach, this approach is more memory-consuming but should be faster due to the single-pass approach. This is the motivation to compare those two approaches in more detail.

Furthermore, it is interesting to test whether combining iterative and multi-centroid approaches could further improve performance. Namely, the multi-centroid and multi-pass approach would consist of the first step with one pass of multi-centroid training and then passing several times through the training data to fine-tune the exact centroid values.  

\subsection{Weighted learning approach}
%ONLINE HD 
Following the data-driven learning rate idea from~\cite{imani_adapthd_2019}, in~\cite{hernandez-cano_real-time_2021}, the multi-pass approach is replaced with single-pass to reduce the training costs. More specifically, as shown in Fig.~\ref{fig:diffApproachesSchematic}D., a naive accumulation of equally important samples is replaced by using the weighting approach before adding the current vector to the prototype vectors. The weight is defined by the similarity of the current vector to the current prototype vectors; the higher the similarity, the lower the weight. This approach identifies the most dominating patterns and lowers model saturation by them. In~\cite{hernandez-cano_real-time_2021}, authors compared this so called "OnlineHD" approach in performance to the previously described iterative approach. Comparable accuracy was achieved on several different classification problems. 

This approach should be as memory-consuming as the traditional one-centroid single-pass, but might be more time-consuming. This is due to the need to update continuously and normalize the prototype vector after adding each training sample (and not at the end of the pass as in other approaches). It is interesting to analyze the performance of this approach and compare it to the two previously proposed methods for preventing the under-representation of less common patterns.

% \begin{figure*}[]
%     \centering
%     \vspace{2mm}
%     \includegraphics[width=\linewidth]{Images/RawSignlExamples.pdf}
%     \caption{\small{Raw signal showing several seizures from subjects 2 and 4 of the CHB-MIT database. Only the first 4 channels are shown.)}} 
%     \label{fig:rasData}
%     \vspace{-4mm}
% \end{figure*}

\section{Experimental Setup}
\label{Sec: Experimental_setup}

Epilepsy is a highly challenging and relevant problem for which science and technology have not yet proposed a good wearable solution for monitoring, detecting, and even predicting seizures to increase patients' quality of life. Thus, we focus on testing HD learning strategies on this use-case with the final goal of reaching a performance that could be acceptable by users. 

\subsection{Databases}
\label{Subsec:databases}

CHB-MIT~\cite{shoeb_application_2009, goldberger_ary_l_physiobank_2000} is a widely known and publicly available database that contains long EEG (electrocardiogram) recordings that can be used to test algorithms not only in a balanced manner (same amount of seizure and non-seizure) but also in more realistic data balance. 
It contains scalp EEG recordings from 24 subjects, and includes 183 seizures, with an average of 7.6 $\pm$ 5.8 seizures per subject. To standardize the experiment, we use the 18 channels from the international 10-20 bipolar montage that are common to all patients.

From the raw database, we prepared a dataset that contains ten times more non-seizure data than seizure data to be closer to a more real-life data balance. The main reason to avoid a balanced scenario, which is common in many works in the literature, is that it can lead to a highly overestimated performance, not achievable during the continuous monitoring with a wearable device~\cite{pale_multi-centroid_2021}.

Non-seizure segments were chosen randomly from available non-seizure data, but excluding data 1 min before and 15 min after a seizure. This data might contain ictal patterns and thus make classes less separable. In reality, performance on this data segment would not be very relevant for the subject as seizures would be either detected slightly earlier (which could be interpreted as prediction) or seizure detection would last longer, which is not critical for the patient either. Moreover, neurologists often have difficulties in defining the exact end of a seizure, and thus many databases do not even have labeled end of the seizure.

\subsection{Feature extraction and mapping to HD vectors}
\label{Subsec:FeaturesUsed}
In this work, we use 46 features as used in~\cite{pale_multi-centroid_2021} to make it comparable, as we use the same database and share one of the approaches. Most of the features were based on~\cite{zanetti_robust_2020} which contains eight frequency spectrum features, 27 entropy-based features and mean amplitude value. Entropy-based features contain sample, permutation, Renyi, Shannon, and Tsallis entropies. Frequency-domain features present power spectral density and the relative power in the five common brain wave frequency bands; delta: [0.5-4] Hz, theta: [4-8] Hz, alpha: [8-12] Hz, beta: [12-30] Hz, gamma: [30-45] Hz, and a low-frequency component ([0-0.5] Hz). These features are considered medically relevant for detecting seizures~\cite{teplan_fundamental_2002}.

% Next, for each feature, its value $HDV_{Val}$ and its index vector $HDV_{ChFeat}$ are bound (XOR), to get $HDV_{ValFeat}$ vectors. Finally, to get a final HD vector representing each $W_{len}$, we bundle (sum and round) $HDV_{ValFeat}$ vectors of all features and channels, as shown in Fig.~\ref{fig:workflowSchematic}. In this approach, we do not distinguish between channels and treat them all equally important. 

\subsection{Validation}

Training and evaluation are performed in a personalized manner. This is due to the subject-specific nature of epileptic seizures and their signal patterns. More specifically, for each subject and seizure, data is preprocessed to contain 10 times more randomly selected non-seizure data and saved to an individual file. Then leave-one-seizure-out cross-validation is performed, where HD models are trained on all but one file (containing one seizure each). Final performances reported per subject are the average of all cross-validation iterations. 

\subsubsection{Performance Evaluation}
\label{Subsec:PerfEvaluation}
We quantify prediction performance with two measures to increase the interpretability of performance, as proposed in~\cite{pale_systematic_2021}. This is an ongoing discussion that  started with~\cite{ziyabari_objective_2019} and~\cite{shah_validation_2020}.
Here, more specifically, we measure performance on the 1) episode level and on the 2) seizure duration level. %As illustrated in Fig.~\ref{fig:perfMetricsIllustration}, 
The episode level performance cares if seizure episodes are correctly detected, but not necessarily that the full duration of seizure and the start and end time points are correctly classified. It treats predictions and true labels as blocks of 1's and 0's and evaluates on a level of these blocks. 
The duration level performance cares about the correct prediction of each time-point, also meaning that seizures need to be correctly classified during their whole duration. This metric can sometimes be challenging to use due to the difficulty in labeling the true start and end of a seizure. % and existence of inter-ictal periods, 
Indeed, it might not be possible to achieve 100\% accuracy. Thus, we use both metrics to give us a better insight into the operation of the proposed algorithms.  

For both levels, sensitivity, precision, and F1 score are calculated. Sensitivity or true positive rate $TPR$ is calculated as $TP/(TP+FN)$, while precision or positive predictive value or $PPV$ is calculated as $TP/(TP+FP)$. F1 score is calculated as the harmonic mean of sensitivity and precision: $2*TPR*PPV/(TPR+PPV)$. Finally, to have a single measure for easier comparison of methods, we calculate the mean value of the F1 score for episodes ($F1E$) and duration($F1D$) as $F1DEmean$. %the geometric mean value of the F1 score for episodes ($F1E$) and duration($F1D$) as $F1DEgmean=sqrt(F1D*F1E)$.

\subsubsection{Label post-processing}
\label{Subsec:LabelPostprocessing}
The raw predictions from the classifier can show unrealistic behavior for dynamics of epileptic seizures (e.g., seizures lasting only a few seconds or separate seizures that are a few seconds apart). Thus, we utilize time information to perform label post-processing, which consists of two steps. In the first step, time information of the signal is exploited to smooth the predictions by going through the predicted labels with a moving average window of a certain size $SW_{len}$ (5s) and then performing majority voting. In the second step, seizures closer than 30s are merged together into one seizure.

\subsubsection{Statistical analysis}
\label{Subsec:statAnalysis}
Due to the high performance variability between subjects, we perform statistical analyses to compare different strategies. We compare each learning approach with the traditional single-pass one-centroid approach using the Wilcoxon statistical test. It compares the performance of two paired groups, and we report the p-value. 

Finally, the code and data required to reproduce the presented results are available online as open-source\footnote{\label{note1}https://c4science.ch/source/LearningImprovForHDcomputingOnEpilepsy/}.

\begin{figure}[]
    %\vspace{2mm}
    \centering
    \includegraphics[width=0.95\linewidth]{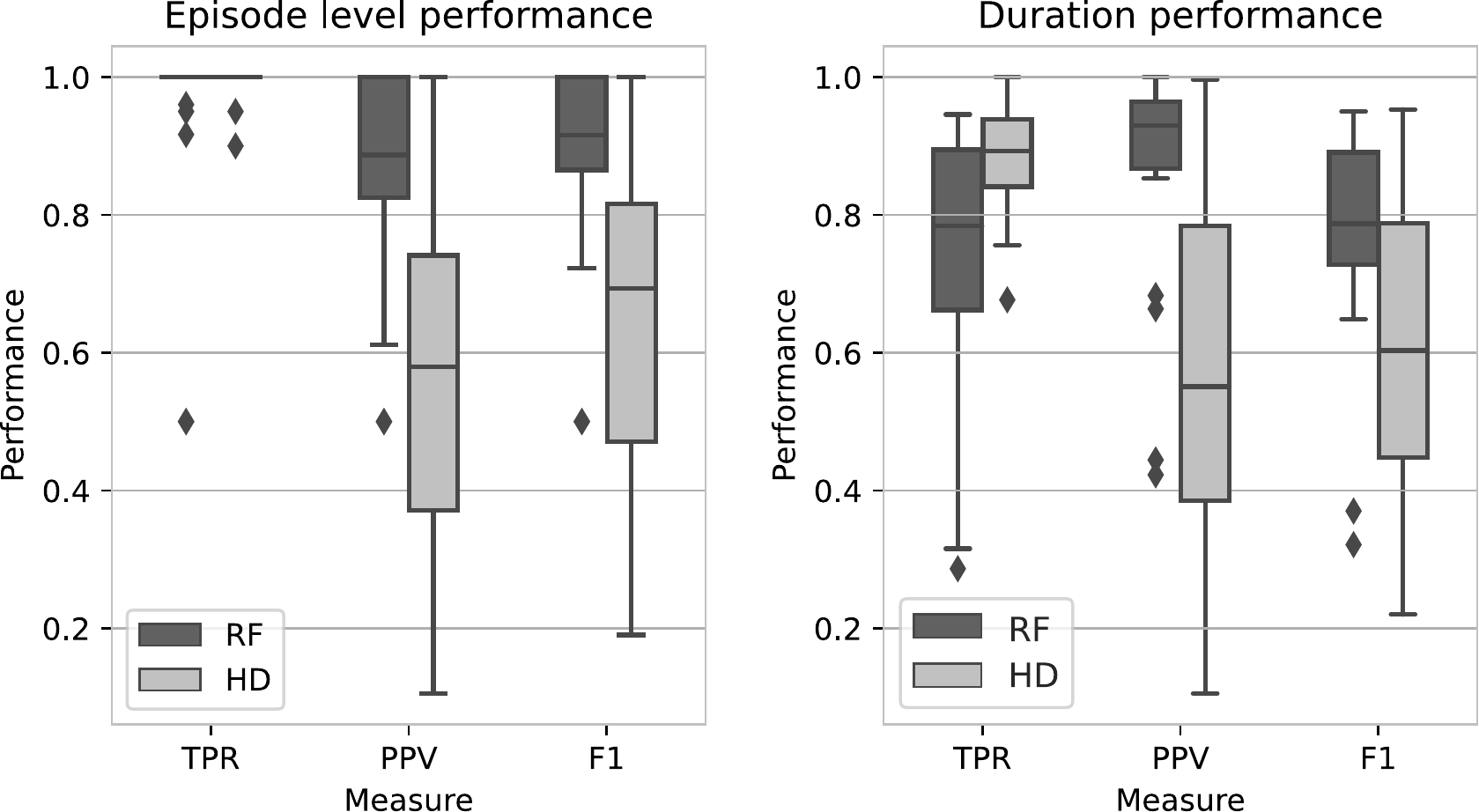}
    \vspace*{-2mm}
    \caption{\small{Performance of random forest as a benchmark for HD computing performance. Performance is shown for both episode and duration level, including sensitivity, precision and F1 score values. In the end, the \textit{mean} value of F1 for episodes and duration is calculated.  }} 
    \label{fig:RFvsHD}
    %\vspace{-2mm}
\end{figure}

\section{Experimental Results}
 
\subsection{Standard HD computing learning  }
\label{Subsec:traditional}

Standard, traditional HD computing learning is single-pass single-centroid per class, where each sample is equally important. As explained previously, it consists of simple summing up (and normalizing in the end) all vectors coming from the same class to get prototype vectors of the class.

Fig.~\ref{fig:RFvsHD} shows a performance comparison of the standard HD computing approach with Random Forest performance as a benchmark. The random forest contained up to 100 trees and was trained on the same dataset, with identical preparation, train-test split, and post-processing as the HD computing approach. 
Here we see that HD computing is performing significantly worse than the random forest approach. TPR or sensitivity stays equally good with the HD approach, meaning that all seizures are usually detected, but a significant drop is perceived in PPV (precision), which means that many false positives occur when using HD computing. For F1 score for episodes the drop in the mean performance of all subjects is 25.7\%, while for F1 duration score is 16.9\%. %The most significant drop is in the precision both for episodes and duration, meaning that there are many false-positive predictions that remain even after label post-processing. 
Thus, in the next experiments, we test if improved learning strategies of HD learning can help resolve this problem and reach the performance of random forest.

\begin{figure}[]
    %\vspace{2mm}
    \centering
    \includegraphics[width=0.95\linewidth]{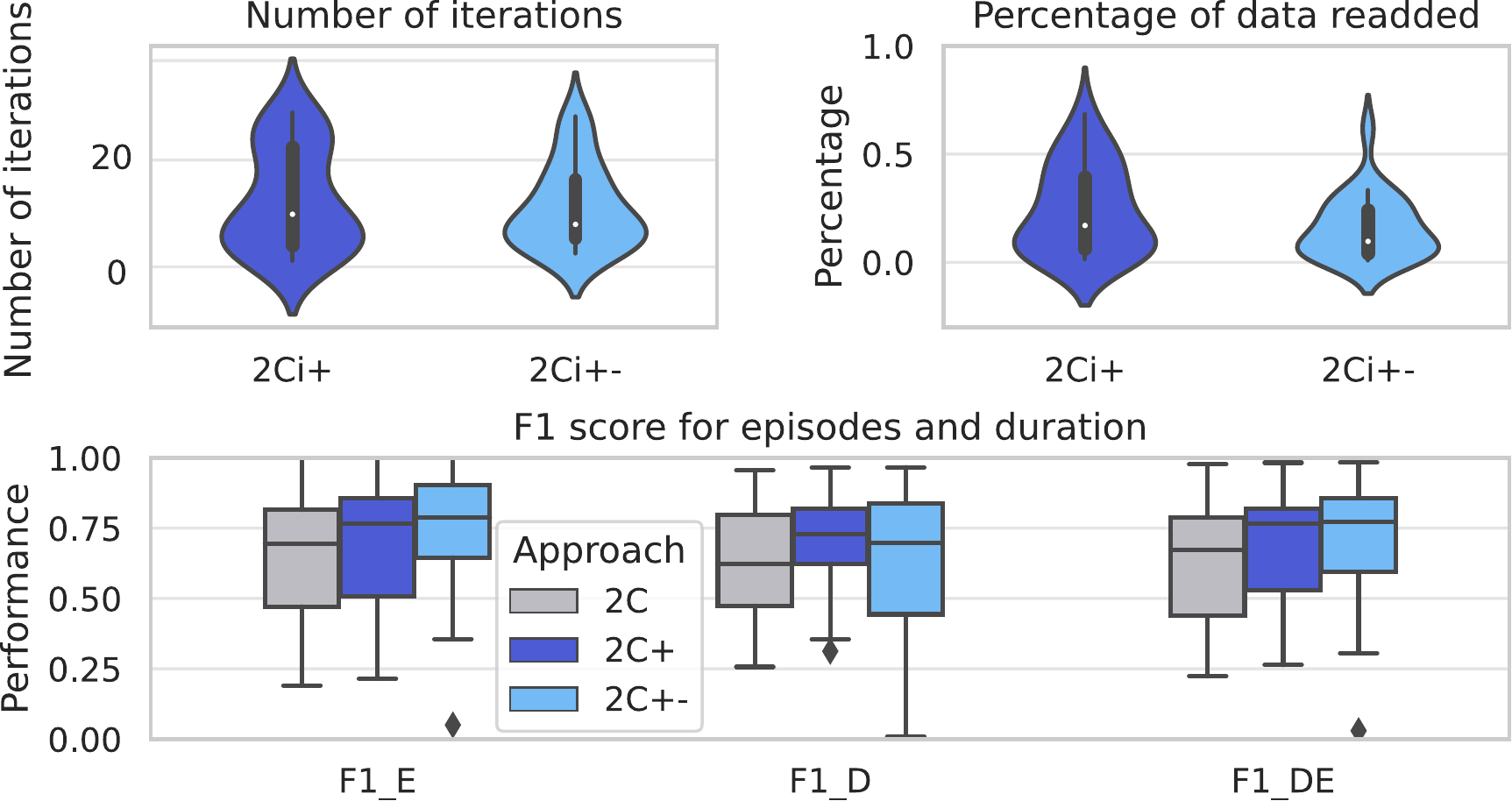}
    \vspace*{-2mm}
    \caption{\small{Iterative (multi-pass) approach compared to single-pass approach. The average number of iterations, percentage of data that had to be re-added and performances on the test set are shown. }} 
    \label{fig:itterativeResults}
    %\vspace{-2mm}
\end{figure}

\subsection{Multi-pass (iterative) learning  }
\label{Subsec:iterative}

In Fig.~\ref{fig:itterativeResults}, two versions of the multi-pass approach are compared with the standard single-pass approach. In one approach, when the current data window would be wrongly classified, it was just added to the correct class again ($2C_+$), whereas in another approach, it was also subtracted from the wrong class ($2C_{+-}$). 
The average number of iterations, percentage of re-added data, and prediction performance are analyzed and shown as a distribution over all subjects. It can be observed that on average, it took around 13 passes for $2C_+$ and 12 passes for $2C_{+-}$ to get a stable performance value. The amount of data that had to be re-added on average was 24\% for $2C_+$ and 16\% for $2C_{+-}$. Interestingly, for $2C_{+-}$ model, the variability between patients is smaller than for the $2C_+$ model.   
Concerning the performance, the F1 score for episodes ($F1_E$) and for duration ($F1_D$) as well as their mean ($F1_{DE}$) is shown. Both multi-pass ($2C_+$ and $2C_{+-}$) approaches have significantly higher performance when compared to single-pass ($2C$) approach ($p=1.63e^{-5}$ for $2C_+$ and $p=2.6e^{-2}$ for $2C_{+-}$ for mean of $F1_E$ and $F1_D$). There is no significant difference in $F1_{DE}$ performance between between using either the $2C_+$ or $2C_{+-}$ approach ($p=0.74$).  
%There is no significant difference between the two multi-pass approaches on the test set, whereas, for the train set, the $2C_{+-}$ approach reaches higher performance. This might be due to over-fitting on a training set. 

\subsection{Multi-centroid learning  }
\label{Subsec:multiCentroidResults}

\begin{figure}[]
    %\vspace{2mm}
    \centering
    \includegraphics[width=0.95\linewidth]{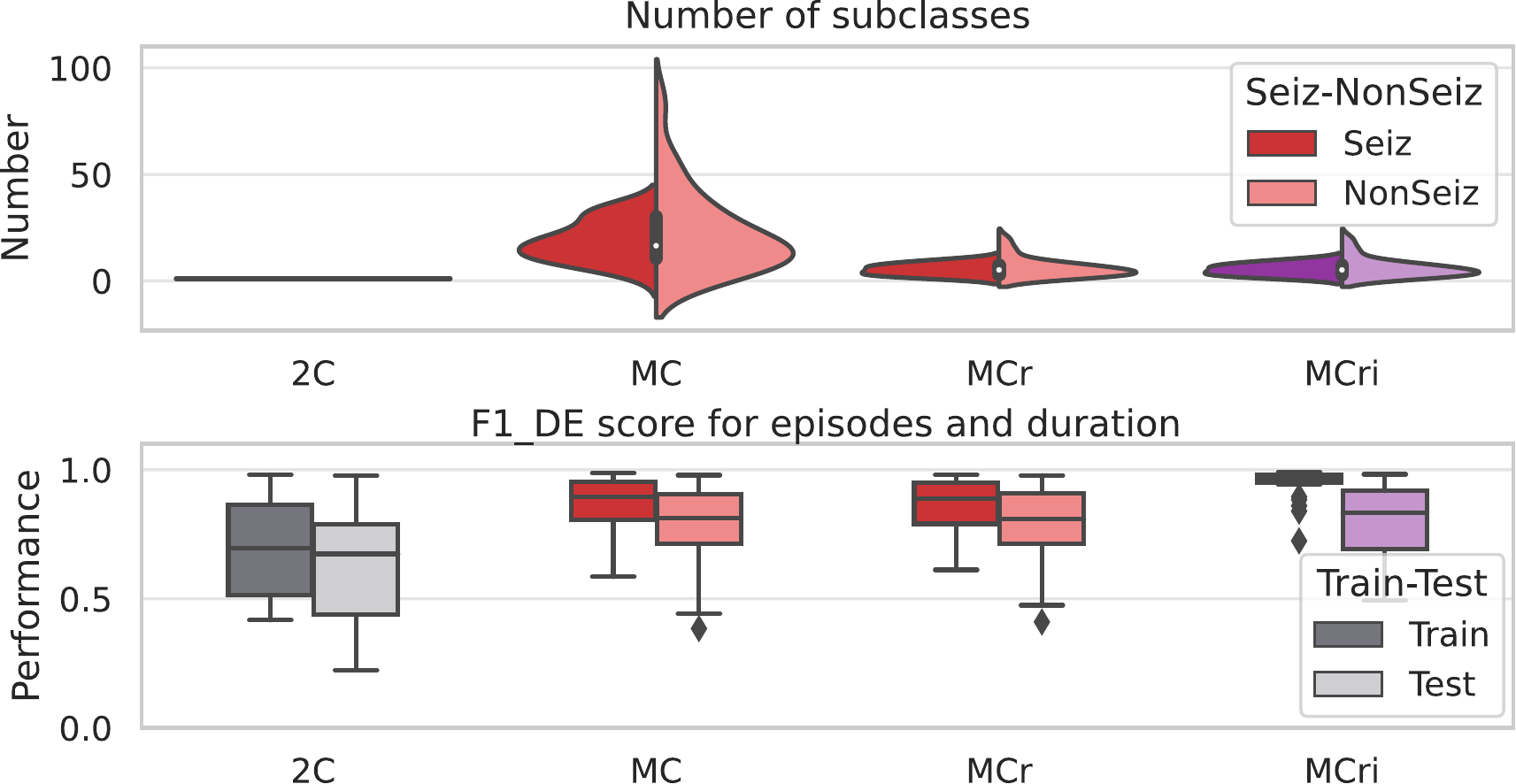}
    \vspace*{-2mm}
    \caption{\small{Multi-centroid learning approach is compared to traditional single-centroid approach. The average number of centroids per seizure and non-seizure as well as performance is shown.}} 
    \label{fig:multiCentroidResults}
    %\vspace{-2mm}
\end{figure}

Fig.~\ref{fig:multiCentroidResults} shows the performance of several versions of a multi-centroid approach when compared to a traditional single-centroid one. As described in~\cite{pale_multi-centroid_2021} first step of the multi-centroid approach is to allow an unlimited creation of sub-classes ($MC$), after which the essential step of removing unnecessary sub-classes follows. This can be done either by simply removing the least common centroids of each class ($MCr$) or clustering ($MCc$) least common centroids with the closest same class ones. Here, results for only $MCr$ approach are showed as they were slightly better and it is lighter to implement. 

Results show that the first step of $MC$ approach creates, in average for all subjects, 18 sub-classes for seizure and 22 for non-seizure. The approach of removing least populated classes ($MCr$) results in, on average, five seizure and six non-seizure sub-classes. %The approach of clustering the least populated classes ($MCc$) results in slightly bigger number of sub-classes; 7 for seizure and 9 for non-seizure. 
Variability is very high between subjects and is within [6-117] sub-classes (summed for seizure and non-seizure) after the first $MC$ step, and is reduced to range [3-26] after $MCr$ step. 
Going from single-centroid to multi-centroid significantly improves the performance on both train and test sets. More specifically, for the test set the average performance of all subjects is increased from 61\% to 76\% for mean value of F1 score for duration and episodes. 
It is important to note that after the second step of reducing the number of sub-classes, despite a significant reduction in the number of sub-classes, the performance is not significantly reduced for the $MCr$ approach. %for neither $MCr$ or $MCc$ approach. 

Next, we tested whether performing additional steps of iterative learning on centroids decided after $MCr$ %or $MCc$
can lead to additional performance improvements. Thus, the number of centroids (and their initial structure) was fixed based on the previous step, but then by passing several times through the training dataset, we allowed slight fine-tuning of the centroids. Although additional performance improvement can be obtained in the training set, but not in the test set. We believe this is due to overfitting. %, and testing on dataset with even more non-seizure data would be interesting (but was not possible with current database). 

\subsection{Weighted learning  }
\label{Subsec:weightedResults}
\begin{figure}[]
    %\vspace{2mm}
    \centering
    \includegraphics[width=0.95\linewidth]{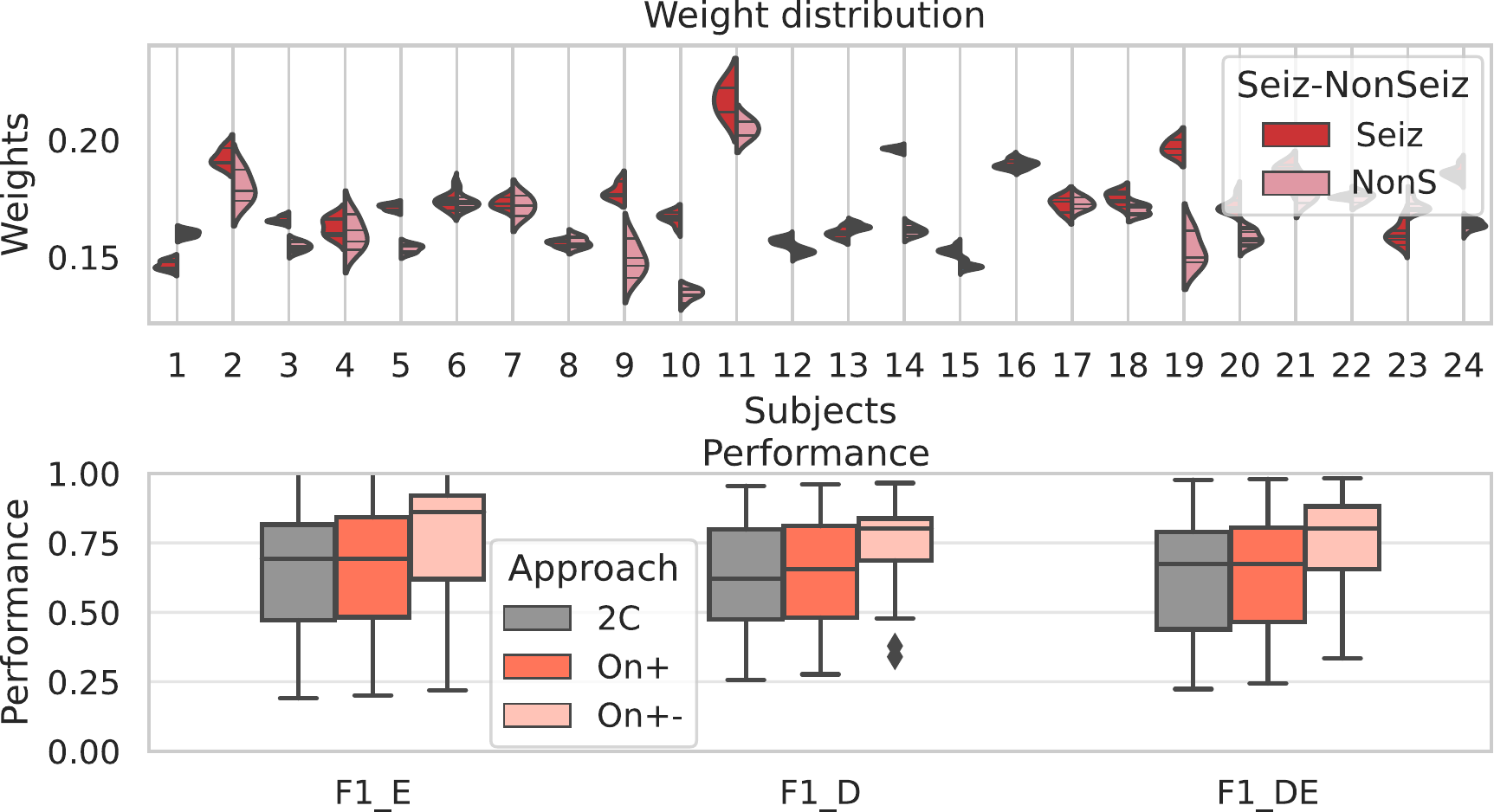}
    \vspace*{-2mm}
    \caption{\small{Weighted learning approach is compared to the traditional single-centroid approach. The distribution of the weights for seizure and non-seizure is shown for each subject on the upper plot. The lower plot shows the performance when comparing two weighted approaches with the traditional (non-weighted) approach. }} 
    \label{fig:weightedResults}
    %\vspace{-2mm}
\end{figure}

 The weighted approach, also called "OnlineHD" ~\cite{hernandez-cano_real-time_2021}, is an alternative to the multi-pass and multi-centroid approaches. % It enhances less common patterns by multiplying them with a higher weight, instead of adding them several times (as in multi-pass) or creating the separate cluster (as in multi-cluster). 
In Fig.~\ref{fig:weightedResults}, the distribution of the weights for seizure and non-seizure is shown. The average values of weights are between 0.1 and 0.2, meaning that between 80\% and 90\% of the bits in vectors were identical when adding a new sample. For some subjects, such as subjects 5, 9, 10, 14, 19 and 24, values are similar for both seizure and non-seizure, while for some subjects, seizure values are larger than for non-seizure, meaning that they were bringing more novelty. For a few subjects (e.g., for subj 1 and 23), it was the opposite, i.e., non-seizure data brought more novelty. 

Finally, when comparing performance, we compared two versions of weighted learning. The first one where the weighed sample is added only to the correct class ($On_+$) and and a second one, similarly to multi-pass, where it is also subtracted from the wrong class if it would be wrongly predicted ($On_{+-}$). 
Performance results in Fig.~\ref{fig:weightedResults} show that just weighted adding ($On_+$) does not significantly improve results on the test set, but that adding and subtracting does help significantly.

\begin{figure}[]
    %\vspace{2mm}
    \centering
    \includegraphics[width=0.95\linewidth]{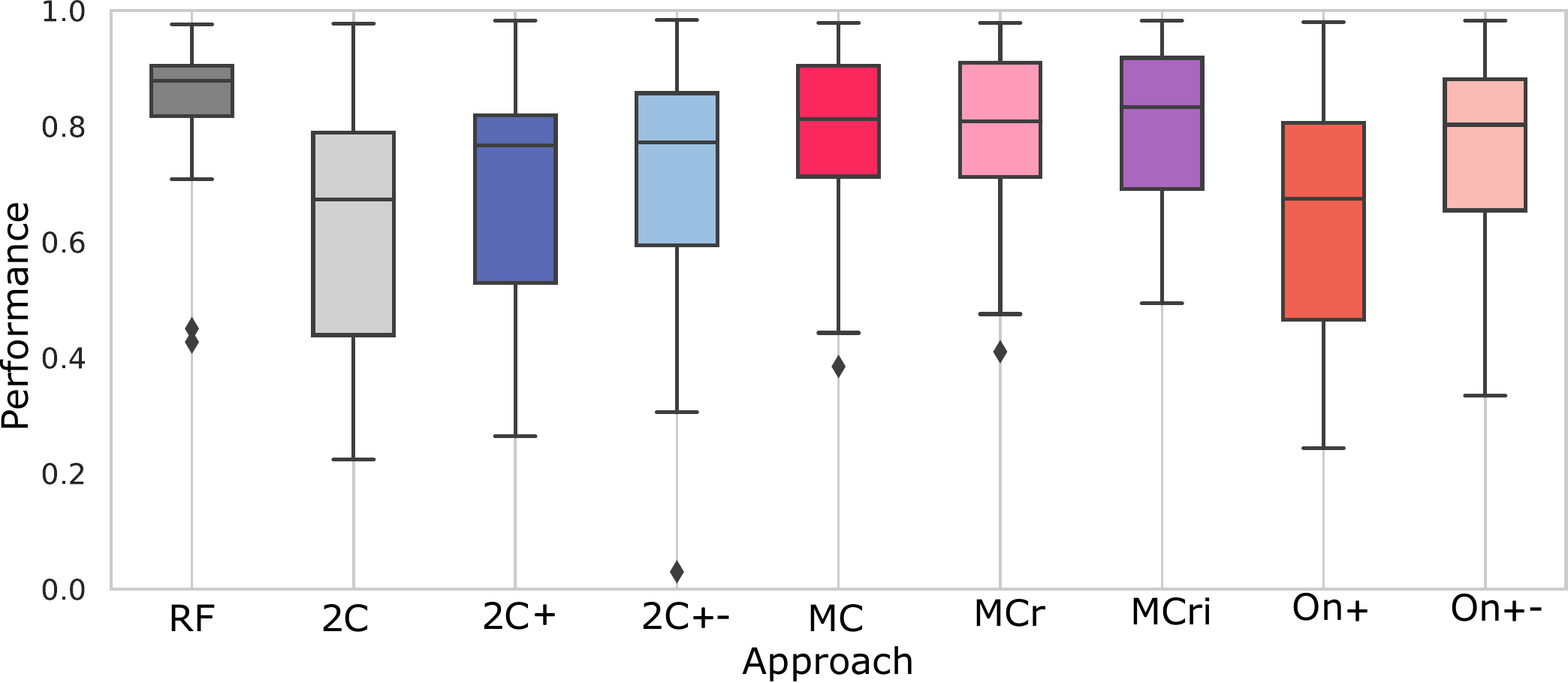}
    \vspace*{-2mm}
    \caption{\small{All learning approaches tested in this work are compared together regarding seizure detection performance. Only one measure combining all together is shown (mean of F1 score for episodes and duration). Results on the test set are shown, where boxplots represent a distribution of performance over all subjects. }} 
    \label{fig:allApproachesResults}
    %\vspace{-2mm}
\end{figure}

\subsection{Comparison of all tested learning strategies}
\label{Subsec:allResultsPerformance}
Finally, in Fig.~\ref{fig:allApproachesResults}, we compare the seizure detection performance for all the studied strategies. The mean of F1 score for duration and episodes is shown as a distribution over all subjects. 
Again, the performance of standard HD computing ($2C$)  with single-pass and single-centroid leads to a significantly lower performance and higher inter-subject variability than the random forest ($RF$) approach. 
Performing iterative (multi-pass) learning improves the performance both for training and test sets, but a larger improvement is possible when both adding the current data vector to the correct class and subtracting it from the wrong class ($2C_{+-}$) than only adding it ($2C_+$), as shown previously in the literature. 

The multi-centroid approach ($MC$, $MCr$%, $MCc$
) also leads to a significantly improved performance (with respect to $2C$) but not yet achieving the level of random forest. If centroids are then fine-tuned through several passes of multi-pass learning ($MCri$% and $MCci$
), performance is improved even more. In that case, there is no significant difference between multi-centroid multi-pass approach with removal of less common classes ($MCri$) with respect to random forest ($p=0.18$). 

The "OnlineHD" approach also leads to a significant improvement in performance, but only when using the approach of adding and subtracting vectors from correct and wrong classes, respectively. However, it did not reach the level of no significant difference with random forest ($p=0.013$). 

%Finally, when comparing the two best approaches, multi-centroid with multi-pass ($MCri$) and "onlineHD" ($On_{+-}$), $MCri$ is slightly better on the train test (not shown here) but not on the test set. This might mean that $MCri$ is over-fitting on the training set. On the test set, they both perform equally well, with no significant difference from the random forest. 

These results show that with additional improvements to standard HD computing, it is possible to achieve a performance as good as with random forests for the seizure detection problem. Nevertheless, Fig.~\ref{fig:allApproachesResults} highlights that variance between subjects is high, especially for HD, and there is space for further improvements of HD performance. %Also, in the future, these approaches could be tested on different epilepsy datasets, with more subjects and also with an even bigger disbalance in the amount of non-seizure and seizure data. 

\subsection{Memory and computational requirements }
\label{Subsec:allResultsCompAndMemory}

\begin{figure}[]
    %\vspace{2mm}
    \centering
    \includegraphics[width=0.95\linewidth]{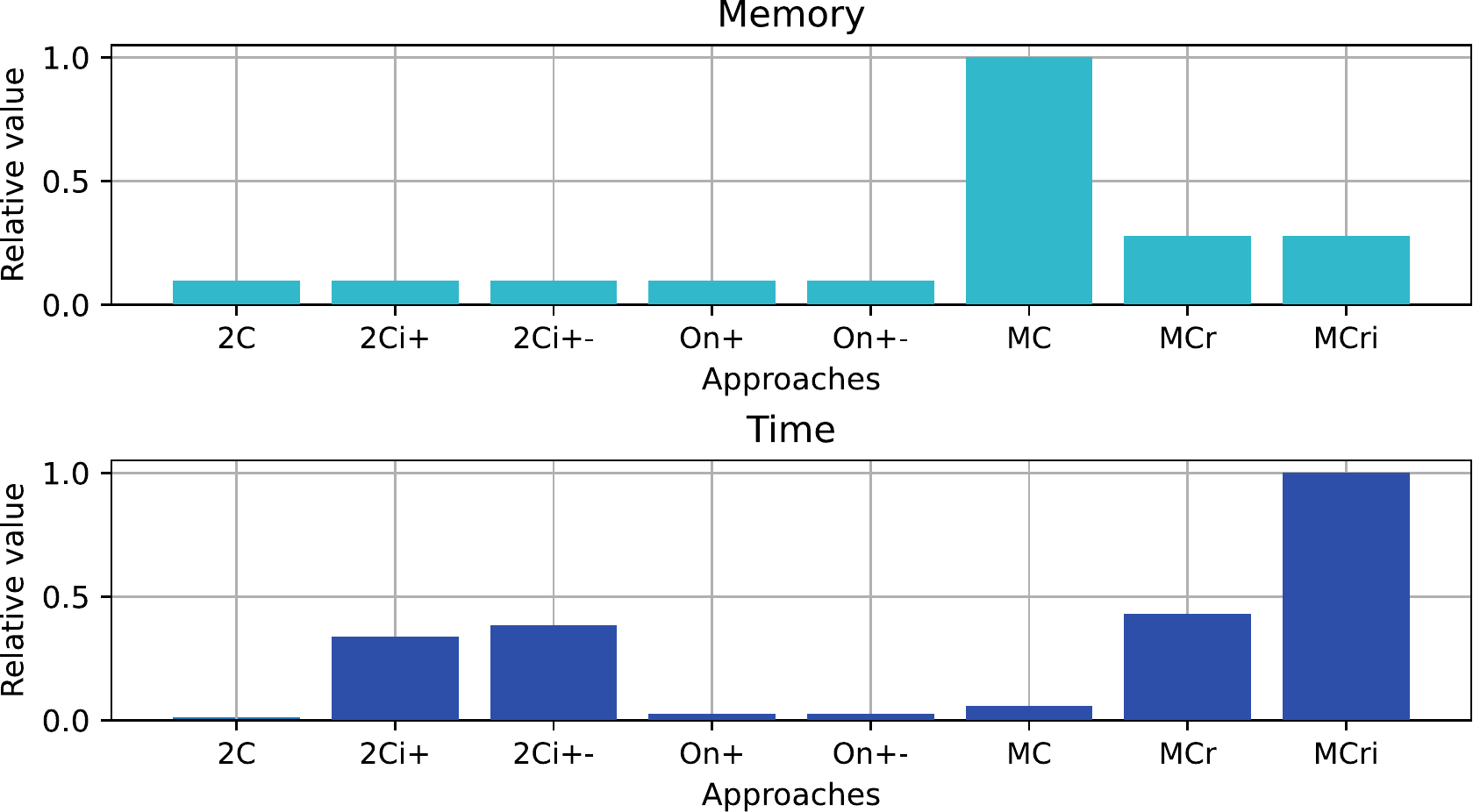}
    \vspace*{-2mm}
    \caption{\small{Comparison of different learning approaches with respect to the time needed for calculation of prototype vectors and memory required for storing them. }} 
    \label{fig:timeAndMemory}
    %\vspace{-2mm}
\end{figure}

Finally, we analyse also the memory and computational complexity of the different learning strategies. Memory is calculated as the memory needed to store all the prototype vectors. In Fig.~\ref{fig:timeAndMemory}, the relative amount of memory is shown to compare different approaches. Multi-centroid learning ($MC$) obviously requires the largest amount of memory, but this is reduced to less than half after optimization steps of removing unnecessary centroids in the second step of the algorithm ($MCr$%and $MCc$
). This is still up to 5 times more than the standard single-centroid "onlineHD" ($On+$ or $On_{+-}$) approach. 

In order to quantify computational complexity we analysed the time needed to perform each type of training using the python implementation%\footnotemark[\ref{note1}]
running on a single-core. We evaluate the relative time between learning strategies. Time was measured as an average over all subjects and all cross-validation iterations. As expected, iterative, multi-pass learning ($2C_+$ or $2C_{+-}$) is significantly more complex (>30x time for $2C$), due to the need for evaluation of performance after each pass of training. Multi-centroid approach in its first step ($MC$) is not extremely time consuming ($\sim$5x time for $2C$), but the second step ($Mrc$ % or $MCr$
) of optimizing the number of centroids increases the computational complexity (>35x time for $2C$). In the end, multi-centroid and multi-pass ($MCri$ %or $MCci$
) is understandably the most time and computationally intensive ($\sim$85x time for $2C$). Interestingly, "onlineHD" ($On_+$ or $On_{+-}$) learning, even though more complex than single-pass, is significantly more lightweight than the multi-pass or multi-centroid approaches (only $\sim$2x time for $2C$).

\section{Conclusion}
In this work, we have investigated and compared the characteristics of different HD computing strategies in the context of epileptic seizure detection with respect to more established approaches such as random forest. In particular, our results have shown a significantly lower performance of the standard HD approach compared to the random forest one, mainly due to many false-positive seizure detections. Moreover, enhancing this standard HD approach by using multi-pass, multi-centroid learning and their combination improves this performance. However, only their combination reaches a performance level comparable to the random forest. 
Then, we tested a single-pass training approach with sample importance ("OnlineHD") and it significantly improved the performance. Furthermore, this approach is much more memory and computationally friendly when compared to a multi-pass approach, which takes more time to train, or when compared to a multi-centroid approach, which requires more memory. Our analyses and results have proven the applicability of HD computing for real-life epileptic seizure detection.

\printbibliography
%\bibliographystyle{IEEEtran}
%\bibliography{references_HDcomp.bib}

\end{document}